\documentclass[runningheads]{llncs}
\usepackage[T1]{fontenc}
\usepackage{cite}
\usepackage{amsmath,amssymb,amsfonts}
\usepackage{algorithmic}
\usepackage{booktabs}

\usepackage{graphicx}
\usepackage{tikz}
\usepackage{textcomp}
\usepackage{xcolor}
\def\BibTeX{{\rm B\kern-.05em{\sc i\kern-.025em b}\kern-.08em
    T\kern-.1667em\lower.7ex\hbox{E}\kern-.125emX}}
\usepackage{graphicx}
\begin{document}
\title{Anomalous Samples for\\ Few-Shot Anomaly Detection\\}
\author{Aymane Abdali\inst{1,2,*} \and
Bartosz Boguslawski\inst{2} \and
Lucas Drumetz\inst{1} \and
Vincent Gripon\inst{1}}
%\authorrunning{A. Abdali et al.}

\institute{IMT Atlantique, UMR CNRS 6285, Lab-STICC, F-29238 Brest, France 
\and
Schneider Electric, Grenoble, France\\
\email{\{*\} aymane.abdali@gmail.com}}
\maketitle              % typeset the header of the contribution
\begin{center}
    \textbf{Preprint. Accepted at INNS DLIA Workshop, IJCNN 2025. To appear in Procedia Computer Science.}
\end{center}

\begin{abstract}
Several anomaly detection and classification methods rely on large amounts of non-anomalous or ``normal'' samples under the assumption that anomalous data is typically harder to acquire. This hypothesis becomes questionable in Few-Shot settings, where as little as one annotated sample can make a significant difference. In this paper, we tackle the question of utilizing anomalous samples in training a model for binary anomaly classification. We propose a methodology that incorporates anomalous samples in a multi-score anomaly detection score leveraging recent Zero-Shot and memory-based techniques. We compare the utility of anomalous samples to that of regular samples and study the benefits and limitations of each. In addition, we propose an augmentation-based validation technique to optimize the aggregation of the different anomaly scores and demonstrate its effectiveness on popular industrial anomaly detection datasets.

\end{abstract}
\section{Introduction}

Visual anomaly detection is the problem of identifying patterns or features in images that deviate from the norm or expected behavior. This field has garnered significant interest because of its widespread applications in areas such as medical imaging~\cite{i4,i5}, industrial inspection~\cite{i1,i2,i3}, and security surveillance~\cite{i6}. Broadly defined, anomaly detection encompasses various problems, such as identifying scratches on the surface of an object or detecting images from a domain different from a predefined one. Anomalies can be physical, like a tear or scratch on an object, or logical, like a misplaced or missing object~\cite{logiconst}.

Focusing on Industrial Anomaly Detection (IAD), this process involves identifying deviations from normal operating conditions in industrial settings. Such anomalies can indicate potential problems such as equipment failures, process inefficiencies, or safety hazards, which makes them crucial for predictive maintenance.

Several paradigms exist for anomaly detection, with unsupervised anomaly detection being the most popular. This approach assumes the availability of a large database of unannotated but mostly normal samples. 

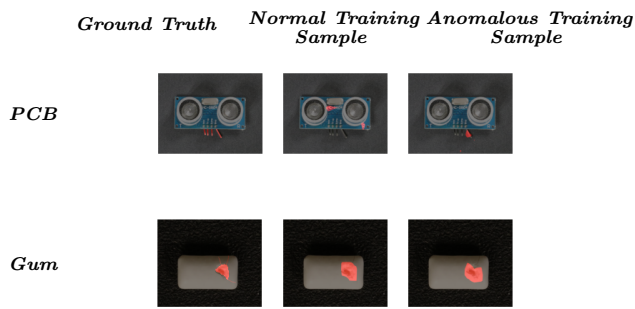
\begin{figure}[t]
    
    \centering
    \begin{tikzpicture}
        % Nodes with images
        \node (image1) at (-1,0) {\includegraphics[width=0.4\textwidth]{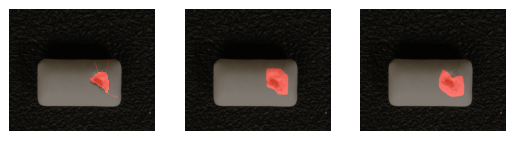}};
        \node (image2) at (-1,2) {\includegraphics[width=0.4\textwidth]{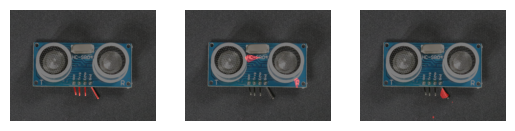}};
        
        \node at (-5,0){\fontsize{7}{10}\selectfont \textbf{\textit{Gum}}};
        \node at (-5,2){\fontsize{7}{10}\selectfont \textbf{\textit{PCB}} };

        \node at (-3.5,3.2){\fontsize{7}{10}\selectfont \textbf{\textit{Ground Truth}}};
        \node[align=center] at (-1,3.25){\fontsize{7}{10}\selectfont  \textbf{\textit{Normal Training}}};
        \node[align=center] at (-1,3){
        \fontsize{7}{10}\selectfont \textbf{\textit{Sample}} };
        \node[align=center] at (1.6,3.25){\fontsize{7}{10}\selectfont  \textbf{\textit{Anomalous Training}}};
        \node[align=center] at (1.6,3){
        \fontsize{7}{10}\selectfont \textbf{\textit{Sample}} };

    \end{tikzpicture}
    \caption{This figure illustrates the potential improvement in anomaly detection segmentation when anomalous samples are included.  We compare the defect segmentation quality when training with a single normal image versus a single anomalous sample. In the PCB class example, the dent representing a defect is not significantly different from the normal reference sample, making it difficult to detect. However, in the gum class, the main damage is obvious enough to be detected. This underscores the benefits of incorporating anomalous samples in training when they are available, while recognizing that they are not applicable in all tasks. Note: while we display the segmentation results in this figure, the paper primarily focuses on the binary classification score.}
    \label{introfig}
\end{figure}

Few-Shot and Zero-Shot anomaly detection models are also gaining attention due to advancements in vision Few-Shot classification. These paradigms operate under the assumption that annotated anomalous samples are difficult to acquire. While this assumption holds in large-data settings, it is less applicable in low-data regimes where it is common to have at least one example of an anomaly. In such cases, even a single anomaly example can enhance performance by capturing specific information about a defect that may be elusive when only looking at normal reference samples, as demonstrated for two classes from the VisA dataset in Figure~\ref{introfig}. Depending on the vision encoder and its pre-training, the latent representation of an emerging anomaly may be more or less discernible using only normal samples. Conversely, a single anomalous sample provides both normal and anomalous regions, which can be used together to enhance the likelihood of detecting the anomaly.
In this paper, we propose:
\begin{itemize}

    \item A methodology that constructs an anomalous memory bank with as few as a single anomalous sample, integrating it with state-of-the-art Zero-Shot and reference memory anomaly scores from previous works.

    \item An efficient, data-free methodology for validating the different anomaly scores, optimizing them into a weighted score through augmentations of the training data.
    
    \item An evaluation of our methodology on Industrial Anomaly Detection (IAD) datasets, namely MVTec, VisA and SDD.
\end{itemize}

\section{Related Work}

In the context of industrial image anomaly detection, many settings are typically considered~\cite{imiad,deepiad}, these include: 
\begin{itemize}
    \item Unsupervised Anomaly Detection where the training set consists exclusively of normal samples.
    \item Fully Supervised Anomaly Detection where the training set consists of both normal and anomalous samples, typically in disparate amounts. The training set can, for instance, be comprised of hundreds of normal samples and less than tens of anomalous samples. 
    \item Few-Shot anomaly detection where we are typically given a training set of only normal samples that doesn't exceed tens of samples.
\end{itemize}
 
 Other paradigms exist to capture the complexities and diversity of real world applications such as Noisy Anomaly Detection which is similar the unsupervised setting with the exception of having a portion of unidentified anomalous amongst the supposed normal training samples~\cite{noisymoe,noisysp}.

\textbf{Unsupervised Anomaly Detection}: An overwhelming amount of the literature focuses on the unsupervised setting where several methods have been put in place~\cite{unsupad}. The most effective solutions are either feature embedding methods that rely on a pre-trained model or reconstruction based methods which self-train encoders and decoders to reconstruct images for anomaly detection. These methods include:
\begin{itemize}
    \item  Teacher-Student models~\cite{effad,asymst}:  the teacher model imparts to the student model the knowledge of extracting normal sample features during training. Then, during inference, the characteristics of normal images extracted from the test set by the teacher network and the student network are comparable, whereas the characteristics of abnormal images extracted from the test set are more distinct.
    \item Memory-based methods~\cite{cpr,padim}: These methods rely on storing normal training sample features in memory, the abnormality of test samples can then be quantified using a distance metric to the reference memory bank. Memory-based methods do not require the loss function for training, and models are constructed quickly. Their performance is ensured by a robust pre-training network and additional memory space, and this type of method is currently the most effective in industrial anomaly detection tasks. They do however require additional memory space to store image features.
    \item Autoencoder models~\cite{autoenc} are the most used reconstruction networks for AD. The key idea behind using autoencoders for anomaly detection is that they are trained to learn compressions of normal data. When they encounter anomalous data, they struggle to reconstruct it accurately, resulting in a higher reconstruction error. This error can be used to identify anomalies.
\end{itemize}

\textbf{Supervised Anomaly Detection}: Fully supervised IAD algorithms make use of the availability of data presenting abnormalities. oftentimes, it only differs from an imbalanced classification task because of the defect segmentation aspect.  Devnet~\cite{devnet} proposes using the deviation loss function to enforce the statistical deviation of all anomalies from normal samples.  DRA~\cite{catchinggrey} incorporates anomalous samples by learning an anomaly classifier head that takes as input a multi-patch representation of the images. Fully supervised anomaly detection is typically associated with better performance but suffers from the high cost associated with anomaly labels in real applications. In addition to being much less popular than their unsupervised counterparts,  these methods are also designed for data abundant scenarios and are not applicable in tasks where only a single example is available. 

\textbf{Few-Shot and Zero-Shot Anomaly Detection}:
    Large pre-trained vision-language models have exhibited impressive Zero-Shot recognition capabilities across various vision tasks~\cite{radford}, including anomaly detection~\cite{jeong}. Notably, CLIP~\cite{clip}, which is pre-trained on billions of image-text pairs, has demonstrated robust generalization abilities in numerous downstream tasks. WinCLIP~\cite{wclip} capitalizes on this by creating a multitude of human-engineered text prompts to harness CLIP's generalizability for Zero-Shot anomaly detection (ZSAD). Additionally, it introduces a multi-scale scoring system based on the similarity to normal reference images, resulting in a more robust anomaly score. Other works, such as the hierarchical semantic feature adaptation framework proposed in~\cite{vislang}, build upon multi-scale embeddings to enhance the detection of subtle and diverse anomalies.

    Prompt Learning is another Vision-Language learning technique similar to Zero-Shot learning where instead of using fixed pre-engineered text prompts, they are constructed from learnable words through a small set of training images. This method has been popularized by CoOp~\cite{coop}. Recent works have adapted it to Zero-Shot or Few-Shot anomaly detection. Such works include AnomalyCLIP~\cite{anclip} which constructs an object-agnostic prompt template of generic abnormality and normality prompts that is learned through a combination of global and local loss functions from auxiliary data, and PromptAD~\cite{promptad} which creates anomaly and normal prompt prototypes using semantic concatenation of normal and anomalous, learnable and fixed, prefixes and suffixes.

    In a similar direction but without relying on textual information, AnomalyDINO~\cite{anodin} introduces a vision-only approach for Few-Shot anomaly detection that leverages self-supervised features from DINOv2 model. It enhances detection performance by employing patch-level representations and a nearest-neighbor scoring strategy, achieving competitive performance even in one-shot scenarios.

However, we believe that neglecting an anomalous sample in Few-Shot settings is not justified as the limitations pertaining the cost to utility ratio in obtaining anomalous samples do not apply similarly to non Few-Shot settings. Most real-world applications will have at least a single anomalous sample available and while it might not be useful in large-scale data scenarios, it can be extremely valuable in a Few-Shot setting.

\section{Methodology}

Our methodology is built upon the WinCLIP model's Zero-Shot and reference memory anomaly score~\cite{wclip} in a way that allows it to learn from anomalous samples. 

We adopt the same Zero-Shot and reference memory scores. In addition to the memory of normal samples, we construct a multi-scale memory for anomalous samples, which we refer to as the anomalous memory bank. This memory bank stores visual encodings of the anomalous regions from anomalous samples. These encodings are then used to compute an anomaly score at inference time based on the test image's similarity to the anomalous encodings.

This approach allows for a single pixel-level annotated anomalous sample to be used to generate both normal and anomalous memories.

In the scope of this study, we aggregate the aforementioned anomaly scores from various sources—pixel anomaly scores derived from the text template, multi-scale encodings, reference memories, and anomalous memories—into a single score per image. We then measure the performance on a binary anomaly classification task. Future work will consider the anomaly segmentation scores.

\subsection{Problem Formulation}

Anomalous samples can originate from a single category, such as a specific type of defect, or from multiple distinct categories, encompassing various types of irregularities. The aim is to learn a model that yields, for every sample $z_i$, an anomaly score $a_{z_{i}}$ that measures the deviation of the sample from what is considered normal.

We assume that we have access to a restricted training set $S_{train}=(z_i^{test},y_i^{train})$ comprising very few samples and a test set $S_{test}=(z_i^{test},y_i^{test})$ for evaluation. The input $z_i$ represents the image, $y_i^{train}$ is the pixel-level annotation of the training images. It is a binary mask sized similarly to the input image where the pixels corresponding to anomalous parts of the image are flagged. For evaluation, we only consider a binary state (normal or anomalous) through $y_i^{test}$ instead of a pixel-level annotation. We use the training set to fit a learner $A$ that maps every image sample to an anomaly score. $A: z \xrightarrow[]{}a_z$. The goal is to achieve a maximal performance on the images contained in the test set.

Similarly to other works in IAD~\cite{wclip,catchinggrey}, we use the AUROC (Area Under the Receiver Operating Characteristic) Curve to evaluate the quality of our binary classifier.

\subsection{Anomaly Score with Positive Samples}

It is common practice for anomaly detection models to use a composed anomaly detection score. WinCLIP combines a zero-shot score and multiple scores for different scales of the images, DRA~\cite{catchinggrey} combines a score from an anomaly classifier head that is trained on patch-wise representation of anomalous samples, a similar score classification head that is trained on pseudo-anomalies generated from normal samples, and a final one that is learned from the residual features relative to the normal samples with the help of a separate classification head. 

\subsubsection{WinCLIP's score}

To compute our anomaly score, we use WinCLIP's manually crafted template of text to encode comprehensive representations of both normal and anomalous states for a given class. These vector staftes are then used to compute a Zero-Shot score for a given image based on a similarity metric that compares the image in question to the normal and anomalous states. Additionally, the anomaly score includes a comparison to a reference memory bank, where encodings of non-anomalous reference images at various scales are stored. During inference, the distance of test images to these encodings is computed to obtain a multi-scale anomaly score.

This multi-scale score represents the deviation of the image from reference training samples at different patch scales. For three scales; a patch scale of size 16x16 pixels, a mid scale of size 32x32 pixels and a large scale of size 48x48 pixels, patch embeddings of the normal images are stored in a  ''reference memory bank`` $R$ . At inference time, for an image $z$, anomaly scores for the patches of the image $F_{ij}$ are computed as the patch's distance from the most similar patch embedding the reference memory bank, then:

$M_{ij} := min_{r \in R}\frac{1}{2}(1-\langle F_{ij},r \rangle).$

WinCLIP's score is computed as such:

$A_{wclip} = \frac{1}{2}(a_{zs}(z)+max_{ij}M_{ij}^W)$
where:
\begin{itemize}
    \item $a_{zs}(x)$ is the Zero-Shot score obtained as the softmax similarity between the image embedding and the two-state average encodings for the anomalous and normal template encodings.
    \item $M_{ij}^W $ is the average, over three scales, of the anomaly segmentation map from the reference memory. The maximum value over all pixels is retained as the image anomaly score.
\end{itemize}

\subsubsection{Multi-Scale Score comprising an anomalous memory}

We consider a fourth larger scale: a 112x112 pixels scale in addition of the three previously mentioned scales.

We also define a multi-scale anomaly memory score that quantifies the similarity of patches of a given image to that of anomalous patches in anomalous images at different scales. To do so, we define the positive anomaly segmentation map at a given scale as:
\begin{equation}
    M^+_{ij} := max_{r \in R}\frac{1}{2}(1+\langle F_{ij},r \rangle)
\end{equation}
This segmentation map is computed for each patch of the image based on its similarity score to the most similar patch from the anomalous memory bank. Consequently, if any of the stored anomalous samples resembles a patch $F_{ij}$ of a test image during inference, the patch is assigned a high anomaly score. Conversely, if all stored anomalous memory patches are significantly different from $F_{ij}$, the anomaly score for that patch is low.

The final anomaly score we use for our method therefore is:
\begin{equation}
    A_{score} = \frac{1}{3}(a_{zs}(z)+max_{ij}M_{ij}^W+max_{ij}M^{W+}_{ij})
\label{anomalyeq}
\end{equation}

where:
\begin{itemize}
    \item $max_{ij}M^{W+}_{ij} = \frac{1}{4}(a_{p_1}+a_{p_2}+a_{p_3}+a_{p_4})$ is the average, over four scales, of the anomaly scores computed from the anomalous memory similarly. Each individual $a_p$ score corresponds the the anomalous memory similarity described by equation~\ref{anomalyeq} for a given scale. The maximum value over all pixels in the anomaly map is retained as the image's anomaly score for each individual scale. We do this instead of aggregating all the scales before the max operation over the pixels in order to simplify the validation of the scales' scores.
    
    \item $max_{ij}M_{ij}^W  = \frac{1}{4}(a_{n_1}+a_{n_2}+a_{n_3}+a_{n_4})$ is the average, over four scales of the anomaly score using the reference memory.

\end{itemize}

\subsection{Validation of Anomaly Scores}
\label{valan}
Our proposed anomaly score is an average of three major scores. The Zero-Shot score, the reference memory score, and the anomalous memory score. Each of the memory scores is a weighted sum of equally weighted scores across four distinct scales. Let $A_{sc} = (a_{zs}, a_{n_1}...a_{n_4},a_{p_1}...a_{p_4})$ be the vector comprising the anomaly scores. 
The score from equation~\ref{anomalyeq} ca be written as $A_{sup} = A_{sc}.E^T$
where $E = (\frac{1}{3} ,\frac{1}{12} ,...\frac{1}{12})$.

\begin{figure}[t]
    
    \centering
    \begin{tikzpicture}
        % Nodes with images
        \node at (-3,5.7){\fontsize{7}{10}\selectfont \textbf{\textit{Anomalous Image}}};
                
        \node at (0.8,5.7){\fontsize{7}{10}\selectfont \textbf{\textit{Defect Segmentation}}};
                
        \node (im1) at (-1,-1) {\includegraphics[width=0.4\textwidth]{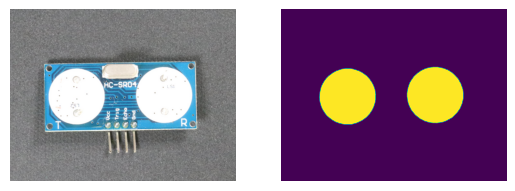}};
        
        \node (im2) at (-1,1.5) {\includegraphics[width=0.4\textwidth]{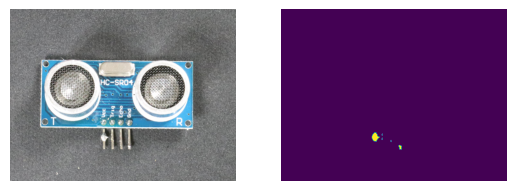}};

        \node (image3) at (-1,4) {\includegraphics[width=0.4\textwidth]{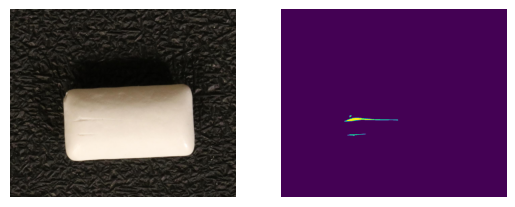}};

    \end{tikzpicture}
    \caption{This figure shows an example of the extreme variation in the size of anomalies in images, whether within the same class or different classes. Some classes, such as printed circuit boards, can have a high variation in the size of the anomaly depending on the anomaly's type. In contrast, a class like chewing gum exhibits much less diversity in how the anomalies arise.}
    \label{figsize}
\end{figure}

\begin{table}[]
\caption{This table presents the performance comparison between training on a normal sample and training on an anomalous sample, using the AUROC score as a metric. For the normal sample training, we use WinCLIP+~\cite{wclip}, and for the anomalous sample, we use our methodology. We present the image AUROC score to determine which setting allows for better discrimination between normal samples and anomalies. The results are presented on three datasets: MVTec and VisA, which contain multiple object classes that we treat as independent objects, and SDD, which contains a single object class. We only ran this experiment on three defined seeds, as the computation of multiple patch-scores for each test image proved to be very compute-intensive.}
\begin{tabular}{lll}
\hline
                       & \begin{tabular}[c]{@{}l@{}}WinCLIP+ \\ (1 Normal Sample)\end{tabular} & \begin{tabular}[c]{@{}l@{}}posWinCLIP+\\ (1 Anomalous Sample)\end{tabular} \\ \cline{1-1}
\textbf{VisA}          &                                                                       &                                                                            \\ \hline
\textit{\textbf{mean}} & 75.90                                                                 & \textbf{81.64}                                                             \\ \hline
candle                 & 90.31                                                                 & \textbf{91.68}                                                             \\
capsules               & \textbf{79.49}                                                        & 77.65                                                                      \\
cashew                 & 91.73                                                                 & \textbf{94.04}                                                             \\
chewinggum             & 95.67                                                                 & \textbf{96.08}                                                             \\
fryum                  & 83.95                                                                 & \textbf{86.52}                                                             \\
macaroni1              & 87.17                                                                 & 86.97                                                                      \\
macaroni2              & \textbf{70.49}                                                        & 68.29                                                                      \\
pcb1                   & 53.51                                                                 & \textbf{91.03}                                                             \\
pcb2                   & 49.51                                                                 & \textbf{52.27}                                                             \\
pcb3                   & 72.05                                                                 & \textbf{72.83}                                                             \\
pcb4                   & 50.48                                                                 & \textbf{74.64}                                                             \\
pipe\_fryum            & 86.71                                                                 & \textbf{87.71}                                                             \\ \cline{1-1}
\textbf{MVTec AD}      &                                                                       &                                                                            \\ \hline
\textit{\textbf{mean}} & 91.51                                                                 & \textbf{92.77}                                                             \\ \hline
bottle                 & 99.33                                                                 & \textbf{99.75}                                                             \\
cable                  & 88.07                                                                 & \textbf{88.97}                                                             \\
capsule                & 63.20                                                                 & \textbf{71.90}                                                             \\
carpet                 & \textbf{99.57}                                                        & 99.36                                                                      \\
grid                   & \textbf{99.32}                                                        & 98.72                                                                      \\
hazelnut               & 95.60                                                                 & \textbf{96.35}                                                             \\
leather                & 99.97                                                                 & \textbf{100}                                                               \\
metal\_nut             & 98.76                                                                 & \textbf{99.24}                                                             \\
pill                   & 90.63                                                                 & \textbf{92.85}                                                             \\
screw                  & 72.62                                                                 & \textbf{73.88}                                                             \\
tile                   & \textbf{100}                                                          & \textbf{100}                                                               \\
toothbrush             & 84.95                                                                 & \textbf{90.51}                                                             \\
transistor             & 88.76                                                                 & \textbf{88.95}                                                             \\
wood                   & 98.25                                                                 & \textbf{99.63}                                                             \\
zipper                 & \textbf{93.55}                                                        & 91.50                                                                      \\ \cline{1-1}
\textbf{SDD}           &                                                                       & \textbf{}                                                                  \\ \hline
                       & 66.3                                                                  & \textbf{68.5}                                                              \\ \hline

\end{tabular}

\label{exp1}
\end{table}

However, each unique object class, depending on the task at hand, can be differently sensitive to each of these scores. Following the relative size of the anomalous region in the image, some scales might be more effective. Larger scales can be more reliable for catching defects in classes where there might be a large variation in defect size in the test set. Figure~\ref{figsize} displays the variety in defect size both inter-class and intra-class. Additionally, the Zero-Shot score depends on a predefined text template. In case of new anomalies arising, that are not well described within the template, this score becomes less efficient. Similarly to the Zero-Shot score, the anomalous memory can fall short if there is a large variety of anomalies. Since this memory creates a score reliant on the similarity to seen anomalies, dealing with anomalies that deviate from what is already seen in training can limit this score's effectiveness. 

This suggests that adapting the scores weights vector $E$ depending on the classes can bring additional performance benefits. We explore this by validating score weights on augmented version of the training images. For a task with a one normal sample for validation and one anomalous sample for training, we generate $m$ augmentations of the images by applying various image augmentation techniques.  We then use a monte-carlo sampling strategy to generate multiple random weights following a probability distribution. For each set of weights, we compute an AUROC validation score on the augmented set. We retain the set of weights that performs the best on our validation set.

\section{Results}

\subsection{Experimental Protocol and Setup}

We conduct our experiments on ViSA, MVTec and SDD datasets. These datasets are originally split into train and test set. The train set contains exclusively normal samples following the traditional paradigm of training only on non anomalous samples. The test set contains different distributions of normal and anomalous samples, where anomalies can be of different classes depending on the object/texture.

We modify these datasets and create splits where randomly drawn anomalous examples serve for training. For each class in these datasets, we consider three runs of anomaly binary classification tasks, each with a different training sample and a up to a hundred test samples (or the remaining number of samples available for that class if it is below a hundred). 

\subsection{Single Sample: Anomalous vs Normal}

\begin{table*}[t]
\centering
\label{testval}
\caption{Decomposition of Anomaly Score in Augmented Training Set on VisA classes}
\begin{tabular}{ll|llll|llll|}
\hline
                                    &                & \multicolumn{4}{l|}{reference memory}                         & \multicolumn{4}{l|}{anomalous memory}                      \\ \hline
\multicolumn{1}{l|}{classes/scores} & text Zero-Shot & patch         & mid           & large         & large+        & patch         & mid           & large         & large+        \\ \hline
\multicolumn{1}{l|}{candle}         & \textbf{83.5}  & 52.4          & \textbf{68.4} & 61.7          & 52.4          & \textbf{79.5} & 63.3          & 66.2          & 59.8          \\
\multicolumn{1}{l|}{capsules}       & \textbf{74.0}  & \textbf{68.7} & 51.9          & 56.2          & 47.2          & 46.2          & \textbf{61.1} & 47.8          & 36.6          \\
\multicolumn{1}{l|}{cashew}         & \textbf{83.3}  & 66.6          & \textbf{68.2} & 52.8          & 62.7          & 63.1          & 51.1          & \textbf{70.1} & 49.3          \\
\multicolumn{1}{l|}{chewinggum}     & \textbf{66.0}  & \textbf{75.3} & 50.7          & 40.9          & 60.6          & \textbf{80.5} & 40.5          & 51.3          & 72.7          \\
\multicolumn{1}{l|}{fryum}          & \textbf{74.9}  & \textbf{80.1} & 57.4          & 62.1          & \textbf{62.2} & 52.1          & 59.4          & 34.7          & 46.6          \\
\multicolumn{1}{l|}{macaroni1}      & 56.3           & 39.7          & 45.8          & 51.8          & \textbf{58.1} & \textbf{68.2} & \textbf{62.3} & 45.0          & 57.7          \\
\multicolumn{1}{l|}{macaroni2}      & \textbf{69.7}  & 65.2          & \textbf{70.8} & 63.9          & \textbf{70.8} & 67.2          & 53.5          & 42.8          & 46.9          \\
\multicolumn{1}{l|}{pcb1}           & \textbf{70.6}  & 65.8          & 45.8          & 60.5          & 61.8          & \textbf{69.0} & 44.8          & 36.1          & \textbf{67.3} \\
\multicolumn{1}{l|}{pcb2}           & 49.8           & \textbf{65.4} & 51.1          & \textbf{64.9} & 59.7          & 34.9          & 54.8          & \textbf{63.8} & 45.0          \\
\multicolumn{1}{l|}{pcb3}           & 47.3           & 51.6          & 44.7          & 48.6          & \textbf{58.1} & 58.0          & \textbf{58.2} & \textbf{59.3} & 51.1          \\
\multicolumn{1}{l|}{pcb4}           & \textbf{76.0}  & 56.6          & 60.3          & \textbf{78.9} & \textbf{67.6} & 46.0          & 46.6          & 49.9          & 53.0          \\
\multicolumn{1}{l|}{pipe\_fryum}    & \textbf{79.1}  & \textbf{79.8} & 66.6          & 68.7          & 62.5          & \textbf{95.8} & 52.7          & 73.6          & 68.0          \\ \hline
\end{tabular}
\label{exp2bis}
\end{table*}

\begin{table*}[t]
\centering
\label{testval}
\caption{Decomposition of Anomaly Score in Test Set of VisA classes}
\begin{tabular}{ll|llll|llll|}
\hline
                                    &                & \multicolumn{4}{l|}{reference memory}                            & \multicolumn{4}{l|}{anomalous memory}      \\ \hline
\multicolumn{1}{l|}{classes/scores} & text Zero-Shot & patch          & mid           & large          & large+         & patch         & mid   & large & large+        \\ \hline
\multicolumn{1}{l|}{candle}         & \textbf{88.55} & 80.18          & 79.08         & \textbf{81.66} & \textbf{81.37} & 58.49         & 51.70 & 68.30 & 73.67         \\
\multicolumn{1}{l|}{capsules}       & 73.67          & \textbf{77.89} & 65.42         & 60.21          & 57.11          & 52.78         & 47.18 & 40.74 & 52.34         \\
\multicolumn{1}{l|}{cashew}         & \textbf{92.9}  & 90.7           & \textbf{61.2} & 66.9           & 59.9           & 62.3          & 47.9  & 50.8  & 50.4          \\
\multicolumn{1}{l|}{chewinggum}     & \textbf{94.3}  & \textbf{94.3}  & 72.9          & 79.7           & \textbf{89.2}  & 61.1          & 33.1  & 36.8  & 39.9          \\
\multicolumn{1}{l|}{fryum}          & 72.4           & \textbf{85.5}  & 64.7          & 69.0           & \textbf{78.5}  & 42.0          & 47.4  & 47.5  & 42.9          \\
\multicolumn{1}{l|}{macaroni1}      & \textbf{79.4}  & \textbf{80.5}  & 62.9          & 58.3           & \textbf{69.7}  & 48.9          & 53.8  & 47.5  & 42.6          \\
\multicolumn{1}{l|}{macaroni2}      & \textbf{64.5}  & \textbf{78.7}  & 56.0          & 55.9           & 60.2           & \textbf{67.3} & 53.9  & 58.8  & 46.9          \\
\multicolumn{1}{l|}{pcb1}           & 72.0           & 64.4           & \textbf{88.6} & 78.2           & 79.8           & \textbf{84.0} & 53.7  & 66.7  & \textbf{81.7} \\
\multicolumn{1}{l|}{pcb2}           & 40.4           & 59.5           & \textbf{65.6} & 58.0           & 65.6           & 44.2          & 49.2  & 56.0  & 50.5          \\
\multicolumn{1}{l|}{pcb3}           & \textbf{68.5}  & \textbf{74.0}  & 51.6          & \textbf{61.4}  & 57.1           & 41.6          & 46.5  & 47.9  & 34.6          \\
\multicolumn{1}{l|}{pcb4}           & \textbf{72.6}  & 40.6           & 50.4          & \textbf{67.9}  & 64.1           & 56.9          & 59.2  & 44.2  & \textbf{77.6} \\
\multicolumn{1}{l|}{pipe\_fryum}    & 72.3           & \textbf{96.9}  & \textbf{87.8} & \textbf{90.9}  & 83.8           & 56.3          & 33.8  & 45.9  & 36.9          \\ \hline
\end{tabular}
\label{exp2}
\end{table*}

In this first experiment the main objective is to quantify the usefulness of anomalous samples in training with respect to our methodology. We compare the WinCLIP+ score to our score that modifies it to include a memory bank for anomalous samples and that also considers an additional larger scale. Table~\ref{exp1} shows that our proposed score yields better results, on most of classes in all three datasets. In VisA, a 5.5\% increase in score can be observed. This is largely due to a significant increase in the scores of specific categories. For the printed circuit board classes, respectively ``pcb1" and `pcb4",  we see a 37\% and 24\% boost in performance when training on a single anomalous sample. Such an increase can occur for classes where the textual state prompts such as ``scratch on a printed circuit board" fail to align properly with the encoded representations of the anomalous regions in the images. Some other classes however, seem to be more discernible using the normal samples. Several reasons can explain that as well; the first one being that there can be several variations of anomalies within the same class that are very different from one another. In such scenarios, constructing a memory bank for anomalous samples can mislead the model when unseen anomalies arise that are more dissimilar to the seen anomaly than to the normal sample. Another simpler potential explanation is that the text template is sufficient to very accurately catch all potential anomalies in the test set.

\subsection{Scores Validation}

\begin{table*}[t]
\centering
\caption{This table shows the different AUROC performance scores for our binary anomaly classifcation tasks with different ways of weighing the final anomaly score. The ``baseline weights'' represents the weights proposed in equation~\ref{anomalyeq}. `random uniform weights'' are a set of weights that is generated randomly once following a uniform distribution. And ``Monte-Carlo Weights'' refer to the weights from our validation methodology which are either validated on the test set, for an ``oracle score'' that serves to show the potential performance gains if we were to find the perfect weight set, or on our augmented. The 95\% confidence intervals from repeating the task over a hundred runs are also included to ensure statistical significance in the cases of random weights samplings and Monte-Carlo.}

\begin{tabular}{l|l|l|l|}
\hline
 & VisA & MVTec & SDD \\ \hline
Baseline Weights & 81.64 & 92.77 &  68.05\\ \hline
Random Uniform Weights & 81.60 $\pm$ 0.10 & 92.40 $\pm$ 0.05 &  66.80 $\pm$ 0.20\\ \hline
Oracle Weights with Monte-Carlo on Test Set (Normal) & 86.50 $\pm$ 0.09 & 94.83 $\pm$ 0.10 &  74.52 $\pm$ 0.14 \\ \hline
Monte-Carlo on Validation Set (Uniform Sampling) & 82.02 $\pm$ 0.09 & 92.60 $\pm$ 0.05 & 69.20 $\pm$ 0.20\\ \hline
Monte-Carlo on Validation Set (Normal Sampling) & 64.42 $\pm$ 0.60 & 82.41 $\pm$ 0.45 &  48.20 $\pm$ 0.45 \\ \hline
Monte-Carlo on Validation Set (Student-t Sampling) & 64.19 $\pm$ 0.65 & 85.41 $\pm$ 0.45 &  45.21 $\pm$ 0.45  \\ \hline
\end{tabular}

\label{exp3}
\end{table*}

\textbf{Validation on Augmentations of Training Samples:}

We use two samples, an anomalous sample to generate the memory banks and a normal sample exclusively for validation. In order to evaluate the robustness of the anomaly scores, we augment both our samples onto ten images using a combination, of image rotation, image flipping, random distortion and skewing.

We can see in the tables~\ref{exp2bis} and~\ref{exp2} that, for any given class, the top three best performing isolated anomaly scores on the test set and on the augmented set have at least one score in common. This augmented set can serve as a proxy for determining potential scores to weigh more when computing the final anomaly score. We speculate that information on what scores might be more suitable for the task can be found within the training samples. Such an example is the size of the defect proportionally to the full image. Simply put, larger scales seem to be more efficient in finding larger defects. The training samples can also be used to validate the text template; some defects can be accurately described with adequate text prompts while it can be more challenging for others. For example, with the typical pre-trained CLIP encoders, a scratch on wood aligns well with the text corresponding to it, whereas ``contaminations'' on ``capsules'' yield weaker activations. Since the training samples are not used in this Zero-Shot text score, they can yield an indication on whether said score is effective or not. We can therefore leverage the training samples in order to weight our scores differently.

In the experiment presented in Table~\ref{exp3}, we re-use the same anomalous training sample and the same tasks as the previous experiments and we use an additional normal sample exclusively for validation. We augment the two samples onto 10 total images, and we validate the weights on this augmented set. We proceed by generating $N=100$ random weights initializations following multiple distributions, the weights that yield the best score on this validation set are kept. We see from Table~\ref{exp3} that in two out of the three datasets, we obtain small but statistically significant performance boosts compared to using the baseline weight vector $E$ in Section~\ref{valan}. These improved results correspond to the Monte-Carlo sampling of weights 100 times using a uniform distribution, We find that other distributions, such as the normal distribution or Student's t-distribution, result in severe overfitting of the weights. We also see that a single uniformly drawn set of weights, for all datasets, performs very similarly to the baseline weights, highlighting the arbitrariness of those weights. We also include the "Oracle Weights" in the Table, which are derived using the same sampling and validation process, but applied to the test set. These results indicate that there is a greater potential for performance improvement if we can more accurately determine the appropriate weights for different scores.
\section{Conclusion and Perspectives}

In this paper, we present our formulation of a Few-Shot anomaly binary classification problem where we incorporate anomalous images into our training process. We challenge the common notion of completely disregarding anomalous samples simply because they are generally pricier to acquire, as the cost of sample acquisition does not scale the same way in Few-Shot scenarios. Therefore, we propose a learning methodology based on recent Zero-Shot and memory-based techniques that incorporates anomalous samples in the training.

We highlight the unique benefits of using anomalous samples while recognizing the potential challenges they can bring. We demonstrate that, more often than not, a single anomalous sample can guide learning more effectively than a single normal one. Additionally, we propose an anomaly validation score method that combines several types of scores—scores from different image scales, scores from both normal and anomalous images, etc.—and weighs them in a way that is most beneficial to the binary anomaly classification task at hand.

Future work should further investigate how this anomalous memory and validation process affects defect segmentation quality and explore more strategies for weighting different anomaly scores, as it is common practice in anomaly detection to combine multiple scores.

\vspace{12pt}
\color{red}

\end{document}